\documentclass[letterpaper, 10 pt, conference]{ieeeconf}  
\IEEEoverridecommandlockouts
\usepackage{cite}
\usepackage{amsmath,amssymb,amsfonts}
\usepackage{graphicx}
\usepackage{textcomp}
\usepackage{xcolor}
\usepackage{eqexpl}
\usepackage{autobreak}
\usepackage{comment}

\usepackage{stfloats} 
\usepackage{amsmath}

\usepackage[utf8]{inputenc}
\usepackage{textgreek}
\usepackage{amssymb}
\usepackage{float}
\usepackage{svg}
\usepackage{tikz}
\usepackage{pgfplots}
\usepackage{booktabs}
\usepackage{authblk}
\usepackage{mathtools}

\usepackage{algorithm}
\usepackage[noend]{algpseudocode}

\usepackage{tcolorbox}

\graphicspath{{Figures/}}
\pgfplotsset{width=8cm,compat=1.17}

\newlength\imagewidth
\newlength\imagescale

\newcommand{\ra}[1]{\renewcommand{\arraystretch}{#1}}

\def\BibTeX{{\rm B\kern-.05em{\sc i\kern-.025em b}\kern-.08em
    T\kern-.1667em\lower.7ex\hbox{E}\kern-.125emX}}
    
\title{Human-Like Autonomous Driving on Dense Traffic}
\author{Mustafa Yildirim$^{1}$  Saber Fallah$^{1}$
\thanks{$^{1}$Mustafa Yildirim and Saber Fallah are with CAV-Lab, Department of Mechanical Engineering Sciences, University of Surrey,
        {\tt\small \{m.yildirim, s.fallah\}@surrey.ac.uk}}%
}
\date{February 2023}

\begin{document}

\maketitle

\begin{abstract}

This paper proposes a imitation learning model for autonomous driving on highway traffic by mimicking human drivers' driving behaviours. The study utilizes the HighD traffic dataset, which is complex, high-dimensional, and diverse in vehicle variations. Imitation learning is an alternative solution to autonomous highway driving that reduces the sample complexity of learning a challenging task compared to reinforcement learning. However, imitation learning has limitations such as vulnerability to compounding errors in unseen states, poor generalization, and inability to predict outlier driver profiles. To address these issues, the paper proposes  mixture density network behaviour cloning model to manage complex and non-linear relationships between inputs and outputs and make more informed decisions about the vehicle's actions. Additional improvement is using collision penalty based on the GAIL model. The paper includes a simulated driving test to demonstrate the effectiveness of the proposed method  based on real traffic scenarios and provides conclusions on its potential impact on autonomous driving.

\end{abstract}

\section{Introduction}

There is growing interest in autonomous vehicles as a solution to traffic congestion, pollution, and vehicle safety issues but yet no solution is for all traffic conditions. Many studies have been performed to solve these issues individually. However, there is still much work to be done to fully realize the potential of autonomous vehicles as a comprehensive solution to these problems.

It is challenging and expensive to manually design the behaviour of an agent in complex environments such as highway driving, and designing the reward function of a Reinforcement Learning (RL) agent requires many iterations. Deep learning models have a tendency to overfit the data. In overfitting, a model's behaviour does not generalize to new or unknown data because it is very sensitive to the training data. Generally, it is more convenient to demonstrate desired behaviour instead of designing it from scratch. Imitation learning is an alternative solution to autonomous highway driving using real driver behaviour. Imitation learning provides less sample complexity to learn a challenging task compared to RL. It is very successful when it is trained on a large dataset but has some drawbacks since its policy does not differentiate the states different than learned states while testing. Therefore it is vulnerable to compounding errors while visited states differ from learned states\cite{hussein2017imitation}. Additionally, imitation learning cannot predict outlier driver profiles and cause collisions in most driving scenarios. Besides this, behavioural cloning often generalizes poorly, mimicking actions exactly even if they are irrelevant to the ultimate task, and fails to learn the expert actions which are intentional and end goal-oriented. 



An early example of imitation learning shows great success for autonomous driving. Pomerlau et al. studied driving a vehicle by mapping states and actions by observing camera images \cite{pomerleau1988alvinn}. Another study investigated the extracting reward function by implementing apprenticeship learning \cite{abbeel2004apprenticeship} on the highway for learning different driving policies. However, this method is only applied in manually generated simulations, and further work to use them in more complex simulations or real-world data is required if such approaches are to catch up to the level of real driving performance. But yet, this implementation needed to be improved regarding vehicle dynamics being modelled unrealistically.
Various methods have been studied, from rule-based methods \cite{niehaus1994probability,gipps1986model,kesting2007general} to state-of-the-art novel reinforcement learning methods\cite{yildirim2022prediction}. Advanced techniques for decision-making in highway driving have been studied to improve upon rule-based methods. Techniques such as decision trees \cite{quinlan1986induction} and random forests\cite{ho1995random} rely heavily on data and are prone to overfitting. SVMs\cite{liu2019novel} are susceptible to noise, and small changes in data can lead to differing results. Game theory\cite{meng2016dynamic} and fuzzy logic\cite{balal2016binary} are effective in low-traffic conditions, but their complexity increases with the number of vehicles. Monte Carlo Tree Search (MCTS)\cite{silver2010monte} is a promising technique, but only a limited number of search branches \cite{gonzalez2019human} are considered due to computational cost. However, most of these techniques still need to improve with generalization and adaptability in new driving conditions. 

Unforeseeable consequences may occur as a result of traffic vehicles' decisions. In some cases, the next vehicle might maintain its velocity and position but might act differently based on the same distance and acceleration parameters. It is impossible to anticipate all scenarios in which autonomous vehicles might violate safety. This causes uncertainty, a massive challenge for ongoing traffic, and typical neural network structures could not handle this problem. A range of different inputs might be the solution for the same output (avoid collision). This requires a mixture of distribution which can be modelled with Mixture Density Network \cite{bishop1994mixture}.

An autonomous vehicle can make more informed decisions about its actions by using MDNs to predict the probability distribution of other vehicles' future positions based on their current positions and velocity. However, determining the optimal number of mixture components for the model is challenging. Few mixture components might cause the model not to capture the complexity of the expert's behaviour, leading to suboptimal performance. On the other hand, many mixture components cause the model may become overcomplex and difficult to train. An MDN may not be well suited for problems where the expert's behaviour is highly non-stationary since it assumes that the distribution of the expert's behaviour remains constant over time. 

Mixture density networks have been used in various applications\cite{kuutti2022end,arbabi2022planning} related to autonomous driving, such as vehicle control and navigation. One of the main advantages of using MDNs in this field is their ability to handle complex and non-linear relationships between inputs and outputs, which is often the case in autonomous driving scenarios. A study proposed AMDN\cite{kuutti2021adversarial} for the vehicle following problem; compared to our method problem is simplified, and action space is limited with longitudinal actions. But for our case, we have extended action space to control two axis such as longitudinal and lateral.

Inherently, a BC model will be restricted by the data points in the dataset. On the other hand, generative models do not limit their solutions to specific instances found in the training data\cite{fabbri2018d}. Variational autoencoders (VAE) \cite{kingma2013auto} are trained in a semi-supervised manner where networks are learned for only one problem, which limits its performance. As GANs learn a distribution, they can learn a wider variety of possible solutions, which makes them a better fit for unseen driving conditions where BC models fail. On the other hand, GAN networks have vulnerability for mode collapse, and vanishing gradients \cite{arjovsky2017towards}. A GAN is usually expected to produce a variety of outputs. The network learns to model a particular data distribution through subsequent training, producing monotonous results. Mode collapse is a common problem in GANs where the generator produces only a limited number of distinct samples, resulting in a loss of diversity \cite{salimans2016improved}. Compared to minimax-based GANs, Wasserstein GANs \cite{arjovsky2017wasserstein} avoid problems with vanishing gradients. Mixture density generative adversarial networks (MDGANs)\cite{eghbal2019mixture} are less prone to mode collapse because they model a mixture of distributions. 


Generative adversarial imitation learning (GAIL) \cite{ho2016generative} represents a promising approach to learning sequential decision-making for highway driving. Unlike RL, GAIL uses demonstration data by experts (human driving dataset) and learns both the unknown environment’s policy and reward function. GAIL can handle complex problems and generalize well to new situations. Moreover, GAIL presented considerable achievement in complex continuous environments. However, it suffers from the requirement of many interactions during the training. Furthermore, it is very time-consuming in real-world applications, where there need to be more interactions between an agent and the environment to achieve an appropriate model\cite{navidi2021new}.

The findings indicate that recurrent GAIL can effectively replicate real trajectories and exhibit desirable properties. Bhattacharyya et al. \cite{bhattacharyya2018multi} expanded on this by applying GAIL to multi-agent learning in complex driving scenarios. Despite the sound methodology of GAIL, there have been no further studies by either the academic or industry communities.

Traditional imitation learning (IL) methods mainly utilize expert data, which can lead to redundancy due to experts' tendency to adhere to optimal policies and avoid risky state-action pairs. The data dependency of imitation learning algorithms investigated by Ross et al.\cite{ross2011reduction}.They proposed an approach (DAgger) involves training a policy using expert demonstrations, generating new trajectories, and iteratively using the expert to correct the behaviour in these new trajectories. This necessitates real-time access to the expert and potentially results in significant costs. The gathered expert data are typically concentrated around optimal state-action pairs, while hazardous pairs are rarely encountered. This scarcity of data related to dangerous situations can hinder the IL's generalization performance. Another study \cite{wang2019random} addresses undesirable terminal states, such as car crashes in autonomous driving, by applying a terminal reward heuristic to deter an agent from ending an episode outside the estimated boundaries of the expert policy. In another study \cite{lee2020mixgail}, authors proposed an algorithm incorporates negative demonstrations, enhancing the learning process by discouraging the agent from entering hazardous state-action regions. This approach provides a more holistic understanding of the task environment, leading to improved performance and safety in complex scenarios.

This paper focuses on generating an agent that on highway traffic by mimicking other vehicles driving. A few studies have used NGSim\cite{yeo2008oversaturated} traffic dataset, but in this study HighD\cite{highDdataset} dataset has been used. The highD dataset is more complex, high dimensional and large in vehicle variations. It represents a broader dataset, which includes 60 road and driving conditions, where NGSIM uses only one road data compared to HighD. As the data size and variety increase, predicting vehicle driving behaviours becomes much more challenging. Whereas it is easy to represent a small dataset by implementing rule-based methods such as Decision Trees, Random Forests, as the dataset gets bigger, these methods lose accuracy and more advanced structures, such as Neural Networks, are needed to represent the model accurately.

In this paper, we present a proposed method for modelling driving behaviour. We first introduce the dataset and simulation platform in Section II as the basis for our model. Section III explains the methods, including BC, MDN, GAN and GAIL. Specifically, we provide an analysis of the results obtained from our proposed method compared to baseline methods. Section IV presents a comprehensive analysis of our results. Finally, we summarize our findings and discuss potential future work in Section V.

\section{Model}

\subsection{Simulation Platform}

There are numerous open-source platforms available to simulate traffic environments, such as Carla\cite{dosovitskiy2017carla}, AirSim\cite{shah2018airsim}, and Sumo \cite{lopez2018microscopic}. However, it was challenging to implement real traffic data simulation on these platforms. As our goal was to simulate an agent in a  real-world application, these platforms were unsuitable. Therefore, we created a Pygame-based traffic simulation environment as depicted in Figure \ref{fig:highd}. Real traffic data was utilized for training and testing, which was simulated in the Pygame simulation platform.

\subsection{Dataset}

\begin{figure}[h]
\centerline{\includegraphics[scale=0.5]{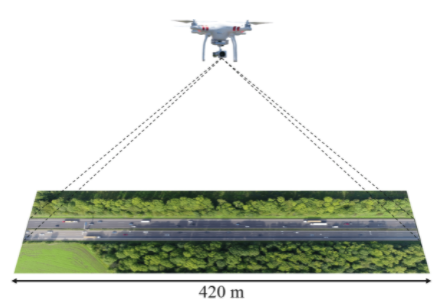}}
\caption{HighD traffic data obtained from \cite{krajewski2018highd}}
\label{fig:highd_drone}
\end{figure}

Studies on highway driving have used two main datasets: NGSim \cite{yeo2008oversaturated} and HighD \cite{krajewski2018highd}. Both datasets contain essential information such as the vehicle's lateral and longitudinal positions, velocity, and acceleration. The HighD dataset was gathered using a drone capturing a 420 m span of the German highway as shown in Figure \ref{fig:highd_drone}, and a wide-angle camera captured vehicle positions, resulting in accurate data in congested traffic. This study used the HighD dataset for training and testing, with 15 tracks from the 60-track dataset for training and another 15 for testing.

The traffic simulation in this study uses real data, resulting in more accurate reflections of real-world scenarios. However, limitations exist as the data doesn't account for other vehicles sensing the EV, leading to unavoidable collisions.

\begin{figure*}[ht]
\centerline{\includegraphics[scale=0.8]{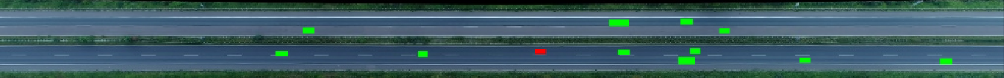}}
\caption{Behavioral cloning model simulation on HighD traffic}
\label{fig:highd}
\end{figure*}

Imitation learning requires demonstration from experts to learn a policy and imitates demonstrated behaviour from dataset. In our case we have 60 different track data which includes more than thousands expert driver behaviour for highway driving. The Behavioral Cloning model, as shown in Figure \ref{fig:highd}, observes surrounding vehicles' velocity and distance and mimics their lateral and longitudinal movement behaviours by controlling acceleration on both axes. This allows the agent vehicle to perform actions such as accelerating, decelerating (braking), and moving left and right. 

Action space is discrete as acceleration varies $a = [a_{min},a_{max}]$ in both axes.

From the traffic, one car is assumed as an expert driver then the distance and velocity of other vehicles are obtained as a state.

\section{Method}

\subsection{Imitation Learning}

In imitation learning, a dataset of demonstrations $D = {(s_{i},a_{i})}$ that consists of pairs of states and actions is used. Using the dataset D, a common optimization- based strategy learns a policy $\pi^{*}$ that satisfies

Imitation learning generates a policy based on examples from a dataset $D = {(s_{i},a_{i})}$ that consists of state and action pairs. An optimization- based method learns a policy $\pi^{*}$ that satisfies

\begin{equation}
\pi^*=\arg \min D(q(\phi), p(\phi))
\end{equation}

where $q(\boldsymbol{\phi})$ is the  expert distribution, $p(\boldsymbol{\phi})$ is learner policy  distribution of the features and $D(q, p)$ is a similarity measure between $q$ and $p$.

Normal Gaussian distribution is not convenient to model if there are multiple correspondences (y) to the same input (x). Bishop proposes a mixture density network \cite{bishop1994mixture} to overcome this problem.

\subsection{Mixture Density Network}
Mixture density network (MDN) is used for predicting the probability distribution of a target variable, rather than just the mean or most likely value. It does this by modeling the target variable as a mixture of different probability distributions, such as Gaussian distributions. This allows the network to better handle complex and non-linear relationships between inputs and outputs, and can lead to more accurate predictions. MDNs are commonly used in applications such as time series forecasting, speech recognition, and image generation. They are trained using the maximum likelihood method, which involves adjusting the parameters of the network to maximize the probability of the observed data.

Unlike Feed Forward Network, Mixture Density Networks are capable of modeling distributions as combinations of parametric distributions:

\begin{equation}
p(y \mid x)=\sum_{i=1}^M \alpha_i(x) \phi\left(y \mid \theta_i\right)
\end{equation}

In the above equation, $y$ is the output, $x$ is the input, $M$ is the number of mixture components, $\alpha_i$ is the mixing coefficient, and $\phi$ is a parametric distribution with parameters $\theta_i$. 

\begin{figure}[H]
\centerline{\includegraphics[scale=0.15]{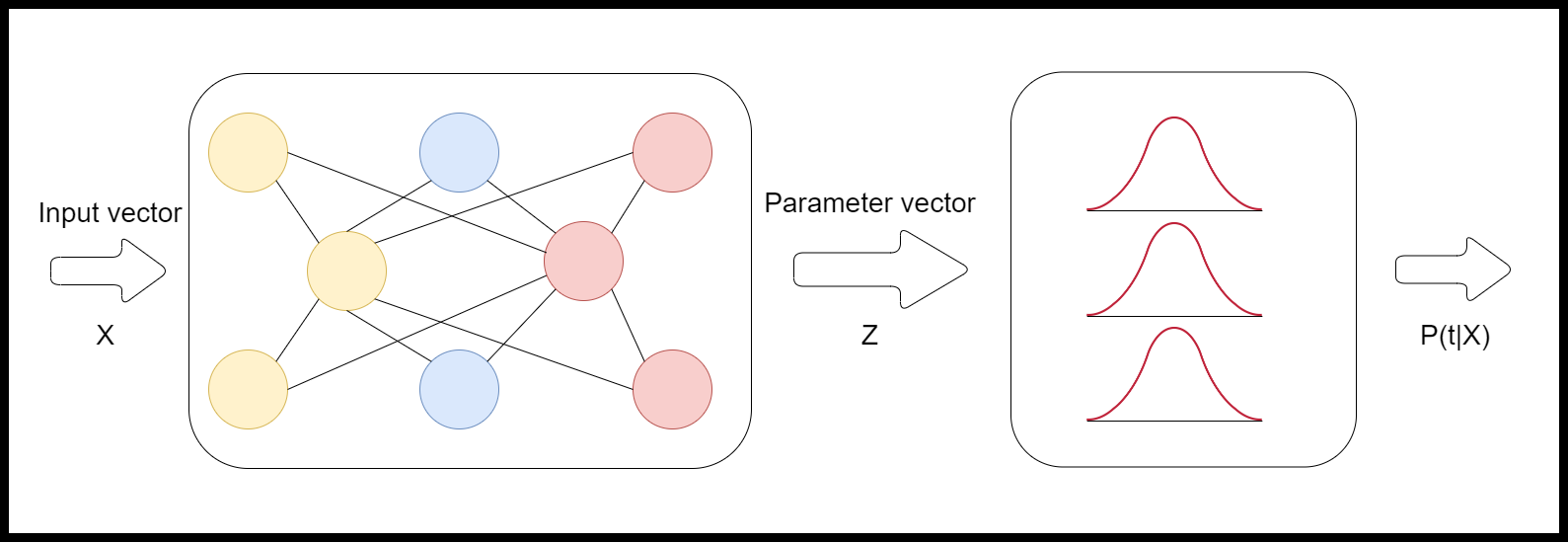}}
\caption{Mixture Density Network Model }
\label{fig:MDN}
\end{figure}

In a Mixture Density Network, the model's parameters are determined by the outputs of a feed-forward neural network. Based on the input vector, the mixture model illustrates the conditional probability density function of the target variables \cite{bishop1994mixture}.

A fully Connected Layer (FCL) supplies the parameters for multiple distributions, which are later on mixed by weights. The resulting conditional probability distribution enables the model of complex patterns for complex highway driving in real-world data.

\subsection{GAN}
We have implemented a Generative Adversarial Network (GAN), like behavioural cloning, to control vehicle acceleration. GAN differs from Behavioral Cloning (BC) in that it consists of two networks: the Generator (G) and the Discriminator (D).
The networks are trained simultaneously, with the Discriminator being trained to maximize the probability of correctly distinguishing whether a sample came from real data or from the Generator. Meanwhile, the Generator is trained to deceive the Discriminator by minimizing $log(1 - D(G(z)))$. The following objective function represents this process:

\begin{equation}
\min _G \max _D \mathbb{E}_{x \sim \mathbb{P}_r}[\log D(x)]+\mathbb{E}_{z \sim \mathbb{P}_z}[\log (1-D(G(z)))]
\end{equation}
where $\boldsymbol{x}$ represents real data, $\boldsymbol{z}$ represents noise samples from a prior distribution $p_z$, $G$ represents the generator network, and $D$ represents the discriminator network.

\subsection{GAIL}

GAIL imitates expert behaviour by formulating the task as a two-player minimax game between a generator and a discriminator network. The former tries to generate samples indistinguishable from the expert's behaviour, while the latter tries to distinguish the generated samples from the expert's behaviour. The training process continues until the generator produces samples that are indistinguishable from the expert's behaviour, at which point it has learned to imitate the expert.

The generator is never exposed to real-world training examples, only the discriminator. Therefore it is able to minimize problems associated with translating expert demonstrations into the target agent's domain by taking advantage of this situation.

\begin{equation}
\begin{aligned}
\max _\psi \min _\theta V(\theta, \psi)= & \underset{(s, a) \sim \mathcal{X}_E}{\mathbb{E}}\left[\log D_\psi(s, a)\right]+ \\
& \underset{(s, a) \sim \mathcal{X}_\theta}{\mathbb{E}}\left[\log \left(1-D_\psi(s, a)\right)\right] .
\end{aligned}
\end{equation}

while reward function can be extracted as surrogate:
\begin{equation}
\tilde{r}\left(s_t, a_t ; \psi\right)=-\log \left(1-D_\psi\left(s_t, a_t\right)\right)
\end{equation}

Proximal Policy Optimization (PPO) mitigates the impact of noisy policy gradients by employing a clipping mechanism that restricts significant policy updates. Discriminators can serve as reward functions for RL, assessing whether the behaviour is similar to the expert.

\begin{tcolorbox}[fonttitle=\bfseries, title=Generative Adversarial Imitation  Learning]
\begin{algorithmic}
\State Expert trajectories $\tau_{e} \sim \pi_{\text {expert }}$ where $\tau_{e}=\left(s_{1}, a_{1}, s_{2}, a_{2}, \ldots\right)$
\State Initialize params $\theta$ of policy $\pi_{\theta}$ and params $w$ of discriminator $d_{w}$

\For{number of iteration}
    
    \State Update discriminator parameters:
    \State $\delta_{w}=\sum_{(s, a) \in \tau_{e}} \nabla_{\mathrm{W}} \log d_{w}(s, a)+\sum_{s, a \sim \pi_{\theta}(a \mid s)} \nabla_{w} \log \left(1-d_{w}(s, a)\right)$
    \State $w \leftarrow w+\alpha_{w} \delta_{w}$
    
    \State Update policy parameters with PPO:
    \begin{align*}
    &\operatorname{Cost}\left(s_{0}, a_{0}\right)=\sum_{s, a \mid s_{0}, a_{0}, \pi_{\theta}} \log \left(1-d_{w}(s, a)\right) \\
    \delta_{\theta} &= \left[\sum_{s, a \mid \pi_{\theta}} \nabla_{\theta} \log \pi_{\theta}(a \mid s) \operatorname{Cost}(s, a)\right] \\
    &\qquad -\lambda \nabla_{\theta} H\left(\pi_{\theta}\right)\\
    &\theta \leftarrow \theta-\alpha_{\theta} \delta_{\theta}
    \end{align*}

\EndFor
\end{algorithmic}
\end{tcolorbox}

\begin{tcolorbox}[fonttitle=\bfseries, title=Augmented Intrinsic Reward GAIL]
\begin{algorithmic}
\State Expert trajectories $\tau_{e} \sim \pi_{\text {expert }}$ where $\tau_{e}=\left(s_{1}, a_{1}, s_{2}, a_{2}, \ldots\right)$
\State Initialize params $\theta$ of policy $\pi_{\theta}$ and params $w$ of discriminator $d_{w}$

\For{number of iteration}
    
    \State Update discriminator parameters:
    \State $\delta_{w}=\sum_{(s, a) \in \tau_{e}} \nabla_{\mathrm{W}} \log d_{w}(s, a)+\sum_{s, a \sim \pi_{\theta}(a \mid s)} \nabla_{w} \log \left(1-d_{w}(s, a)\right)$
    \State $w \leftarrow w+\alpha_{w} \delta_{w}$
    
    \State Update policy parameters with PPO and Penalty:
    \begin{align*}
    &\operatorname{Cost}\left(s_{0}, a_{0}\right)=\sum_{s, a \mid s_{0}, a_{0}, \pi_{\theta}} \log \left(1-d_{w}(s, a)\right) \\
    \delta_{\theta} &= \left[\sum_{s, a \mid \pi_{\theta}} \nabla_{\theta} \log \pi_{\theta}(a \mid s) \operatorname{Cost}(s, a)\right] \\
    &\qquad -\lambda \nabla_{\theta} H\left(\pi_{\theta}\right) \\
    &\qquad +  \sum_{\text{col }} \nabla_{\theta} \log \pi_{\theta}(a \mid s) \cdot \text{penalty}(s, a)\\
    &\theta \leftarrow \theta-\alpha_{\theta} \delta_{\theta}
    \end{align*}

\EndFor
\end{algorithmic}
\end{tcolorbox}

\section{Augmented Intrinsic Reward GAIL}

One drawback of GAIL is its reliance on the discriminator to evaluate the expert's behaviour, which may result in inaccuracies when the expert's behaviour is poorly represented in the dataset. Consequently, critical situations, such as collisions, are not be  captured in the expert's demonstrations since expert demonstrations derived from real life scenarios, hindering the agent's effective learning. To address this limitation, we suggest implementing a penalty directly from the environment for such situations. Furthermore, similar to the DAgger algorithm, we  update the dataset to enhance the learning process. This strategy improves the data quality by adding negative state-action pairs in the dataset, which might lead to an overall improvement in agent performance. By doing so, we can enhance the learning process while still benefiting from both an expert dataset and Reinforcement Learning (RL) on the same platform.

\section{Analysis \& Result }

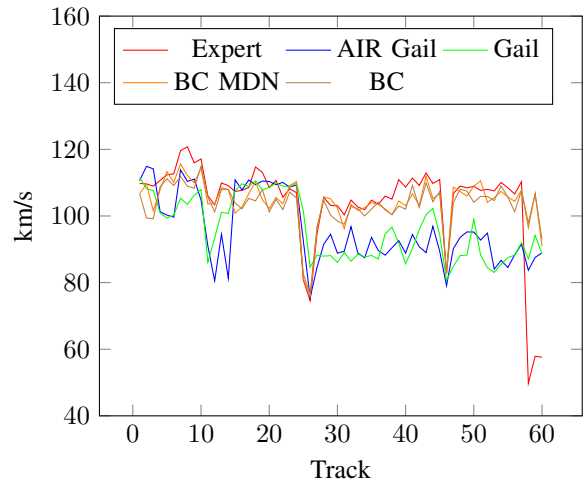
\begin{figure}
    \centering
\begin{tikzpicture}
        \begin{axis}[
            xlabel = {Track},
            ylabel = {km/s},
            ymin = 40,
            ymax = 160,
            legend entries={Expert, AIR Gail,Gail,BC MDN, BC}, 
            legend pos=north west, 
            legend columns=3 
        ]
        \addplot[color=red] table [x=Track, y=xVelocity_kmph, col sep=comma] {average_values.csv};
        \addplot[color=blue] table [x=Track, y=Aug_Gail_vel, col sep=comma] {average_values.csv}; 
        \addplot[color=green] table [x=Track, y=Gail_vel, col sep=comma] {average_values.csv}; 
        \addplot[color=orange] table [x=Track, y=MDN_BC_vel_km, col sep=comma] {average_values.csv}; 
        \addplot[color=brown] table [x=Track, y=BC_vel_km, col sep=comma] {average_values.csv}; 

        \end{axis}
\end{tikzpicture}
    \caption{Average velocity profile of HighD dataset}
    \label{fig:highd_vel}
\end{figure}

\begin{figure}
    \centering
\begin{tikzpicture}
        \begin{axis}[
            xlabel = {Track},
            ylabel = {$\mathrm{m/s^2}$},
            ymax = 5,
            legend entries={Expert, AIR Gail, Gail,BC MDN,BC}, 
            legend pos=north west, 
            legend columns=3 
        ]
        \addplot[color=red] table [x=Track, y=xAcceleration, col sep=comma] {average_values.csv};
        \addplot[color=blue] table [x=Track, y=Aug_Gail_Acc, col sep=comma] {average_values.csv}; 
        \addplot[color=green] table [x=Track, y=Gail_acc, col sep=comma] {average_values.csv}; 
        \addplot[color=orange] table [x=Track, y=MDN_BC_acc, col sep=comma] {average_values.csv}; 
        \addplot[color=brown] table [x=Track, y=BC_acc, col sep=comma] {average_values.csv}; 
        
        \end{axis}
\end{tikzpicture}
    \caption{Average acceleration profile of HighD dataset}
    \label{fig:highd_acc}
\end{figure}
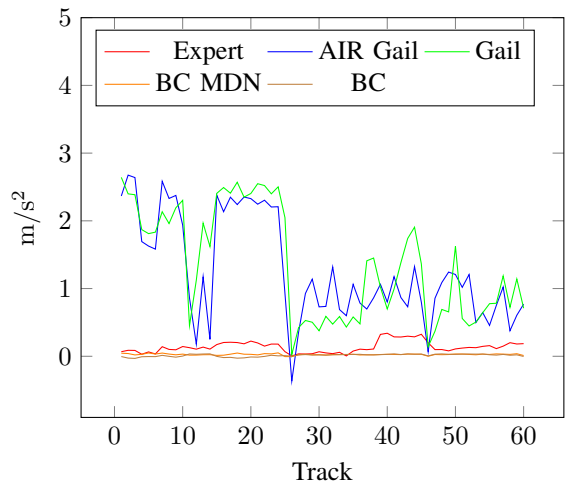

Acceleration and velocity profiles of human drivers are extracted as shown in Figure \ref{fig:highd_vel} and \ref{fig:highd_acc}. The graphs show that human drivers have totally different driving behaviour on every single track. These different behaviours make the imitation learning process even more challenging than usual.

As seen in the Figure \ref{fig:highd_vel}, BC and MDN BC closely align with the velocity and acceleration profiles of the expert drivers. On the other hand, the policies of the GAIL and AIR GAIL models correlate with velocity to some extent. However, their acceleration profiles follow the same trend as velocity, which is markedly different from that of the expert drivers

The GAIL and AIR GAIL models exhibit a unique trait, correlate with velocity to a certain extent, reflecting the velocity changes of the expert drivers. However, these models behave different than experts for acceleration. Instead of reflecting the acceleration that expert drivers would demonstrate in real-world driving, these models align their acceleration profiles with their velocity profiles.
This consistent trend between acceleration and velocity was observed in the GAIL and AIR GAIL models due to these models' underlying policies. They might focus more on velocity than capturing the more nuanced driving strategies involving different acceleration techniques.
Such acceleration behaviour, being consistent with velocity, might result in less realistic driving simulations. Real drivers do not always have a direct correlation between their speed and rate of acceleration. For instance, a real driver might maintain a steady velocity in various road conditions, adjusting acceleration as needed. However, these models might lack this level of subtlety in their driving patterns.
This divergence in acceleration patterns between the expert drivers and the GAIL and AIR GAIL models highlights the need for further refinement in  policy. 

The table represents the performance of various models on a  dense highway driving scenario from the real world. The models include Behavioral Cloning, Behavioral Cloning MDN and GAN, GAIL and AIR-GAIL.  The human column represents the human driver's average values in 60 tracks.

In terms of average collision, the best-performing model is Behavioral Cloning MDN, with an average collision of 10.57. It's followed by Behavioral Cloning at 11.43. The generative models' performance is lower than behaviour cloning models such as  GAIL, which has 34.67 and AIR-GAIL, with 30.35 collisions. The GAN model recorded an average collision of 18.63.

In terms of velocity, human drivers have an average of 104.32 km/h velocity. Notably, the Behavioral Cloning MDN is the closest in matching the human driver's with a velocity of 103.968 km/h, followed closely by Behavioral Cloning at 102.96 km/h. These figures emphasise the potential of these models to mimic expert human performance. On the other hand, while GAIL surpassed the human driver with an unexpected velocity of 122.90 km/h, it's worth noting that this exceeds the expert's speed, which might raise concerns about its practical and safe implementation on highways. Meanwhile, models like GAN and AIR-GAIL were further from the expert influence, recording 98.964 km/h velocities and a considerably lower 90.82 km/h, respectively. Thus, Behavioral Cloning MDN is a promising prospect in pursuing autonomous driving.

The human expert has an acceleration of 0.13 $m/s^2$. In this context, AIR-GAIL emerges remarkably close to this mark, recording an acceleration of 0.69 $m/s^2$. Just behind it, GAIL shows a similar performance with an acceleration of 0.70 $m/s^2$. While more conservative, Behavioural Cloning MDN and Behavioral Cloning are also commendably aligned with the human experts, achieving 0.02 $m/s^2$ and 0.01 $m/s^2$ respectively. Contrarily, with its 2.95 $m/s^2$ acceleration, the GAN model notably surpasses the expert's value, raising potential concerns for its implementation in real-world scenarios considering safety margins. In the ongoing endeavour to mimic human drivers' expertise, AIR-GAIL and GAIL, followed closely by Behavioral Cloning variants, appear to be the closest models for acceleration.

For lane change, 857 human drivers performed 141 lane changes with as average of 0.17 for specific tracks whereas BC performed 8 lane changes, BC MDN 3, GAN 16, GAIL 26 and AIR-GAIL 22.

The results reveal that the models exhibit varied performances concerning collision avoidance, average velocity, and acceleration. Specifically, the GAN model's average velocity and acceleration figures are notably distinct from the other models, emphasizing the unique characteristics of its training or design.

When comparing the models' outcomes against human experts, it's evident that most models haven't similarities to the expert's average velocity. Behavioural Cloning MDN emerges as a prospect, although it still lags slightly behind the human experts. In terms of acceleration, while AIR-GAIL and GAIL demonstrate promise by closely aligning with the human expert, some models, like the GAN, diverge considerably from the expert's metrics. These comparisons highlight the challenges and complexities of replicating human driving expertise through imitation learning models.

\begin{table*}[!htb]\centering
\caption{ Comparison of control policies: Behaviour Cloning with Feed-Forward Network and with  Mixture Density Network, GAN , GAIL and AIR-GAIL}
\ra{1.5}
\begin{tabular}{@{}rccccccccccccccc@{}}\toprule
& \multicolumn{1}{c}{Human}& \multicolumn{1}{c}{BC } &
 &\multicolumn{1}{c}{BC MDN}& &\multicolumn{1}{c}{GAN}&&\multicolumn{1}{c}{GAIL}&&\multicolumn{1}{c}{AIR-GAIL}\\
 \cmidrule{3-3} \cmidrule{5-5} \cmidrule{7-7} \cmidrule{9-9} \cmidrule{11-11}

Collision &- &11.43 &  &   10.57  & &18.63&  &34.67 &&30.35 \\
Velocity km/h &104.32&102.96 &  & 103.968 & &98.964& &122.90&&90.82  \\
Acceleration $m/s^2$ &0.13&0.01  & &  0.02  & &2.95& &  0.70&&0.69\\
Lane Change  &0.17 & 8  & &  3  & &16 & &  26 && 22\\

\bottomrule
\end{tabular}
\label{tab3}
\end{table*}

According to the table, the models have yet to reach the performance levels of human drivers regarding average velocity and acceleration. However, it is worth noting that the models from traffic could not sense our agent driver, which is the main reason for having collision in the scenario. Since other vehicles will act based on our actions in real life, our model may represent the human drivers for highway driving.

\section{Discussion}


Our study compared the performance of three different approaches to autonomous driving on highways: BC, GAN, GAIL and AIR-GAIL. Our results show that mixture density network BC outperformed GAN and GAIL based models in terms of accuracy and stability.
One possible explanation for the lower performance of the GAN and GAIL model is its inherent instability. Generative models are challenging to train and prone to mode collapse, where the generator produces a limited set of outputs that do not cover the entire target distribution. This can lead to a biased or incomplete training dataset, which can, in turn, affect the model's performance. In our study, we observed that the GAN model had a higher prediction variance and was less consistent in its behaviour than the BC model. This suggests that the GAN model suffered from mode collapse or other training instabilities that affected its performance.

Although behavioural cloning has limitations due to the quality and diversity of training data, which can lead to reduced performance in novel or unpredictable situations, its effectiveness can be improved by combining it with a mixture density network. Unlike traditional networks that generate a single output to the entire dataset, MDN uses different kernels for different branches of the mapping\cite{bishop1994mixture}. Due to its improved performance and lower computational requirements, BC with MDN is a more suitable option for highway driving.

One of the disadvantages of using MDNs in autonomous driving is that the training process can be computationally expensive and time-consuming, especially when working with large amounts of data. Additionally, the network parameters need to be carefully chosen and fine-tuned to ensure that the predictions are accurate and consistent.

The Figure \ref{fig:highd_vel} and \ref{fig:highd_acc} represent the human driver's velocity and acceleration values based on the different tracks. It is worth mentioning that acceleration values and velocities differ for each track. That makes our model challenging compared to randomly generated traffic in literature.

In the case of adding one more vehicle to the current traffic, the traffic dataset does not cover all close-distance situations. Adding an additional vehicle is unrealistic to the current traffic environment since traffic already occurred without our agent vehicle. Consequently, the agent often encounters states that the expert did not discover during demonstrations, which generates a more challenging traffic environment. Also, this causes agents to observe unknown states that cause unexpected behaviours and collisions.





Augmented GAIL agent since it gets penalty aswell from collision learns to avoid collision but generate a different behaviour than traffic in the end. It learns to coast on highway while it is less dense or empty but then when vehicle number increase it starts accelerate and continue to navigate as it suppose to as other agents. This coasting behaviour is source of avoid collision by extending time on empty road is interesting behaviour to investigate. As a result of this behaviour, the overall acceleration should be lower than traffic but collision number is less than GAIL.

In conclusion, our study provides evidence that Behaviour Cloning with mixture density network is a more effective approach to highway autonomous driving compared to Generative Adversarial models. While generative models have shown promise in other applications, their inherent instability and tendency for mode collapse make them a challenging choice for autonomous driving tasks. Future research may explore ways to mitigate these issues and improve the performance of generative models for autonomous driving.


\clearpage
\bibliographystyle{unsrt}
\bibliography{main}

\end{document}